\documentclass[11pt]{article}

\usepackage[margin=1in]{geometry}
\usepackage[T1]{fontenc}
\usepackage{times}
\usepackage{amsmath,amssymb,amsthm,mathtools}
\usepackage{booktabs,multirow,array}
\usepackage{graphicx}
\usepackage{xcolor}
\usepackage{xspace}
\usepackage[numbers,sort&compress]{natbib}
\usepackage[colorlinks,citecolor=blue!60!black,linkcolor=blue!60!black,urlcolor=blue!60!black]{hyperref}
\usepackage[capitalise,noabbrev]{cleveref}

\newtheorem{proposition}{Proposition}

\theoremstyle{remark}

\allowdisplaybreaks[1]
\newcommand{\method}{\textsc{ELGO}\xspace}

\newcommand{\nloops}{T}
\newcommand{\params}{\theta}
\newcommand{\initparams}{\theta_0}
\newcommand{\baseparams}{\bar{\theta}}
\newcommand{\refreshint}{K}
\newcommand{\klthresh}{\kappa}
\newcommand{\Ploop}[2]{P^{(#1)}_{#2}}
\newcommand{\student}{\pi_{\params}}
\newcommand{\studentold}{\pi_{\params_{\mathrm{old}}}}
\newcommand{\teacher}{\pi_{\mathrm{T}}}
\newcommand{\lbase}[1]{\ell_{#1}}
\newcommand{\cdw}{w}
\newcommand{\refloop}{r}
\newcommand{\loopdir}{f}

\newcommand{\prompt}{x}
\newcommand{\resp}{y}
\newcommand{\prefix}{s}
\newcommand{\groupsize}{G}
\newcommand{\nresp}{N}
\newcommand{\reward}{R}
\newcommand{\adv}{\hat{A}}
\newcommand{\logratio}{\delta}
\newcommand{\kdcoef}{\beta}

\newcommand{\KL}[2]{\mathrm{KL}\!\left(#1\,\middle\|\,#2\right)}
\newcommand{\E}{\mathbb{E}}

\newcommand{\Var}{\mathrm{Var}}

\DeclareMathOperator{\softmax}{softmax}

\newcommand{\Lgrpo}{\mathcal{L}_{\mathrm{GRPO}}}
\newcommand{\Lkd}{\mathcal{L}_{\mathrm{KD}}}
\newcommand{\Ltotal}{\mathcal{L}}

\title{Beyond the Last Loop:\\Reinforcement Learning with a Teacher\\Extrapolated from the Model Itself}
\author{Linghua Zhang}
\date{}

\begin{document}
\maketitle

\begin{abstract}
Looped language models apply one shared transformer stack several times per token, and every loop can be read out through the shared language-model head. Post-training them with verifiable rewards is hard: the outcome reward carries one bit per response, spreading it over loops trains intermediate loops on tokens sampled from the last loop, and distillation needs either a teacher that ignores the reward or an external, larger model. We observe that a looped model already contains a teacher stronger than its last loop: the next-token distribution extrapolated from the first loop through the last. \method (Extrapolated-Loop Group-relative Optimization) adds to GRPO a dense per-token distillation term towards this distribution. The teacher is computed from an anchor copy of the model, which stays frozen and is replaced by the current model once the student's KL to the teacher falls below a threshold, so the teacher improves with the student without chasing its own extrapolation. The outcome reward splits probability between correct and incorrect responses, and the teacher decides which tokens are preferred inside a response. Between refreshes the objective is KL-regularised RL anchored at the teacher, whose optimum tilts the teacher by the reward. On Ouro-1.4B-Thinking, \method raises the average avg@8 over five math benchmarks from 62.32 with GRPO to 64.20 and is the best compared method on every benchmark, while spreading the reward over loops falls below the base model.

\end{abstract}

\section{Introduction}\label{sec:intro}

Looped language models reuse one set of transformer parameters across several passes per token~\citep{universaltransformer,ouro}.
Ouro~\citep{ouro} applies the same stack $\nloops=4$ times per token and decodes every loop $t$ into a next-token distribution $\Ploop{t}{\params}$ with the shared language-model head; rollouts are sampled from the last loop.
Reinforcement learning with verifiable rewards (RLVR), usually GRPO~\citep{grpo,deepseekr1}, is the main tool for improving their reasoning.

\begin{figure}[t]
  \centering
  \includegraphics[width=\linewidth]{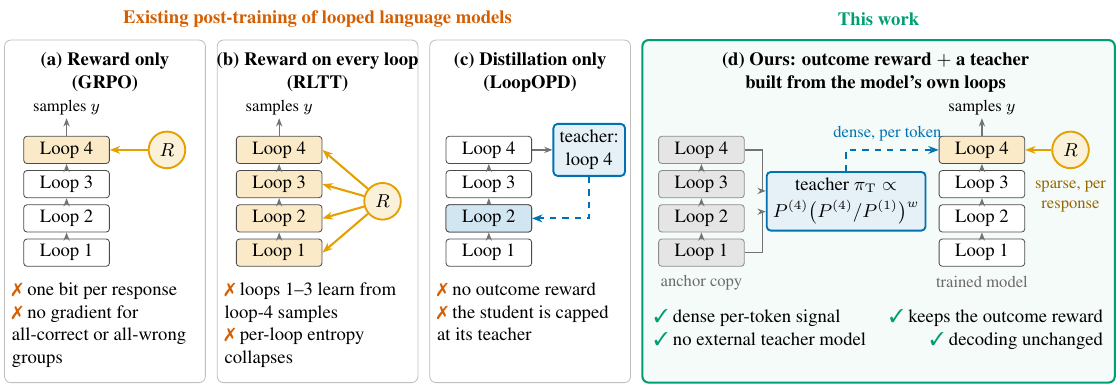}
  \caption{\textbf{Left:} GRPO gets one bit per response; RLTT trains intermediate loops on last-loop samples and their entropy collapses; LoopOPD has no reward. \textbf{Right:} \method extrapolates loop~1 to loop~$\nloops$ of an anchor copy of the model, refreshed when the student has absorbed it, into a dense target for the last loop, next to the outcome reward.}
  \label{fig:motivation}
\end{figure}

Existing methods fall short in three ways (\cref{fig:motivation}, left).
First, reward-only training is sparse: GRPO assigns one advantage to every token of a response, and a group whose responses are all correct or all wrong gives no gradient.
On Ouro-1.4B-Thinking trained on DeepMath problems~\citep{deepmath}, 100 GRPO steps raise avg@8 over five math benchmarks by only 1.17 points (61.15 to 62.32).
Second, RLTT~\citep{rltt} spreads the reward over loops, training intermediate loops on tokens sampled from the last loop.
Running the authors' released code in our setup, loop-1 token entropy falls to 23\% of its initial value and avg@8 drops below the base model on all five benchmarks, by 10.99 points on average (\cref{sec:exp-main,sec:exp-dynamics}).
Third, distillation ignores the reward: LoopOPD's teacher, the model's own terminal loop, knows nothing about correctness and bounds the student, and KDRL~\citep{kdrl} combines distillation with GRPO but needs an external, larger teacher.

Our goal is therefore to add to GRPO a dense per-token signal that comes from the looped model itself, points to a policy better than its last loop, and leaves room for the reward to go beyond it.
Our key observation is that a looped model already carries a teacher stronger than its own last loop.
Cross-loop contrastive decoding (LoopCD)~\citep{loopcd} extrapolates the log-probabilities from loop~1 through loop~$\nloops$, $(1+\cdw)\log\Ploop{\nloops}{}-\cdw\log\Ploop{1}{}$, and without training raises the AIME24 pass@1 of Ouro-1.4B-Thinking from 50.83 to 59.17.
\citet{loopcd} attribute most of the gain to re-ranking rather than to a temperature change, so the gain is not only sharpening.
Training with this teacher raises three challenges.
\textbf{C1:} a teacher recomputed from the current model at every step extrapolates its own extrapolation and has no stable fixed point, while a teacher that never moves is soon absorbed and then only holds the student near the initial model.
\textbf{C2:} the loop signal of a single token is noisy; re-weighting the outcome advantage by it multiplies a noisy signal into a sparse one.
\textbf{C3:} the sparse outcome term and the dense teacher term have different scales, and one can dominate the other.

We propose \method (Extrapolated-Loop Group-relative Optimization), which adds to GRPO a dense distillation term towards a LoopCD teacher that is refreshed by a KL criterion (\cref{fig:motivation}, right).
The teacher $\teacher=\softmax\big((1+\cdw)\lbase{\nloops}-\cdw\lbase{\refloop}\big)$, with $\lbase{t}=\log\Ploop{t}{\baseparams}$, $\cdw=0.5$ and $\refloop=1$, comes from an anchor copy $\baseparams$ of the model. The anchor starts at the initial model and stays frozen; every $\refreshint$ steps, if the student's average KL to the teacher has fallen below a threshold $\klthresh$, it is replaced by the current model (C1).
The dense term is KDRL's k2 estimator of $\KL{\student}{\teacher}$ on rollouts of the student $\student$, the last loop of the trained model, added to the loss, not multiplied into the advantage (C2).
Both terms average over tokens within each response, then over responses, with a constant weight $\kdcoef=0.01$ on the dense term, so $\kdcoef=0$ recovers GRPO exactly (C3).
The outcome reward decides how probability is split between correct and incorrect responses; the teacher supplies a per-token reward that decides which tokens are preferred inside a response.
Between refreshes the objective is KL-regularised RL anchored at the teacher, with optimum $\pi^*(\resp\mid\prompt)\propto\teacher(\resp\mid\prompt)\exp\big(\reward(\prompt,\resp)/\kdcoef\big)$ (\cref{sec:method-analysis}).

Our contributions are as follows.
\begin{itemize}
  \item \textbf{Method.} \method adds to GRPO a dense distillation term towards the LoopCD distribution of an anchor copy of the model that is refreshed when the KL falls below a threshold, designed around C1--C3 (\cref{sec:method}).
  \item \textbf{Analysis.} We derive the closed-form optimum, decompose it into a sparse outcome reward, a dense loop-direction reward and a trust region (\cref{sec:method-analysis}).
  \item \textbf{Experiments.} On Ouro-1.4B-Thinking and five math benchmarks, \method is the best of the base model, GRPO, RLTT and on-policy self-distillation~\citep{opsd} on every benchmark, with an average avg@8 of 64.20 against 62.32 for GRPO and 61.77 for self-distillation; per-loop diagnostics show that RLTT's loss of accuracy comes with a collapse of loop entropy (\cref{sec:exp}).
\end{itemize}

\section{Related Work}\label{sec:related}

\paragraph{Looped language models and their post-training.}
Universal Transformers loop one shared block with adaptive halting~\citep{universaltransformer}.
Ouro~\citep{ouro} pretrains this design on 7.7T tokens; its exploratory RLVR with DAPO and GRPO gave no significant gain over the SFT checkpoint.
RLTT~\citep{rltt} replaces the last-loop log-probability in the policy gradient by a weighted sum over loops with one shared outcome advantage, applied to tokens sampled from the last loop.
LSRL~\citep{lsrl} adds external-grader scores of decoded intermediate states of the Huginn recurrent model to the outcome reward, but gradients flow only through the final-depth output.
LoopOPD~\citep{loopopd} removes the reward: an intermediate loop, on its own samples, minimises the full-vocabulary reverse KL to the last loop of the frozen initial model; its dynamic variant D-LoopOPD replaces this teacher by the stop-gradient last loop of the current parameters as training proceeds.
\method also lets the model teach itself, but its teacher is extrapolated and is refreshed only when the student has absorbed it, its student is the last loop with unchanged rollouts, and it keeps the reward.

\paragraph{On-policy distillation with RL.}
On-policy distillation (OPD) trains a student on its own rollouts towards a teacher's next-token distribution.
KDRL~\citep{kdrl} adds reverse-KL distillation from a fixed external teacher to GRPO as $J_{\mathrm{GRPO}}-\beta D^{k2}_{\mathrm{KL}}$, whose k2 estimator we reuse (\cref{eq:kd}).
ExOPD~\citep{exopd} casts OPD as dense KL-regularised RL with equal reward and KL weights; a reward scale $\lambda>1$ extrapolates the optimum beyond the teacher.
D$^2$-OPD~\citep{d2opd} bounds the extrapolated target's distance from the teacher by a per-prefix KL constraint, and EOPD~\citep{eopd} adds a top-$k$ forward KL at high-entropy teacher positions.
All four use a separate teacher model, and the last three drop the outcome reward.
OPSD~\citep{opsd} avoids a second model by giving the teacher privileged context: the initial model, conditioned on a reference solution, teaches the same model without it through a full-vocabulary divergence on the student's rollouts, again without the outcome reward.
\method needs no privileged context; its teacher comes from the model's own extra loop computation.
Our teacher has the extrapolated form of ExOPD's optimum with $\lambda=1+\cdw$, but it extrapolates between two loops of one copy of the model (loop $\nloops$ over loop $\refloop$) instead of between two separate models, so it needs no second model, and \method keeps the outcome reward on top of it.

\paragraph{Contrastive and extrapolated decoding.}
Contrastive Decoding~\citep{cd} corrects a large expert's logits with a small amateur's, and DoLa~\citep{dola} contrasts the final layer with a premature layer chosen per token.
LoopCD~\citep{loopcd} contrasts loops, decoding from $z_R+\omega(z_R-z_1)$, and attributes most of its gain to the component orthogonal to $z_R$, which re-ranks tokens.
These methods change only decoding, so every test-time token needs the contrast; \method internalises the extrapolated distribution by training and decodes from the last loop alone.

\section{Method}\label{sec:method}

\begin{figure}[t]
  \centering
  \includegraphics[width=\linewidth]{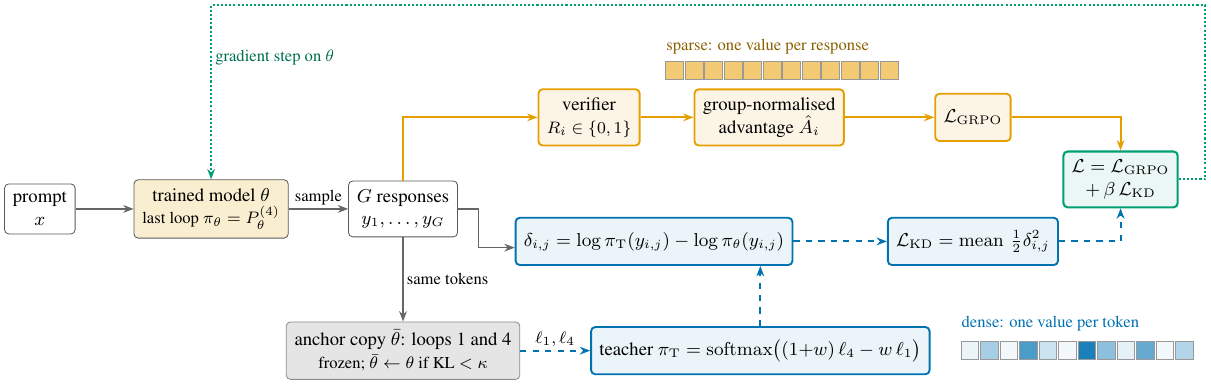}
  \caption{Overview of \method. The anchor copy $\baseparams$ reads out loops $1$ and $4$ on every
  rollout prefix and extrapolates them into the teacher $\teacher$. The anchor starts at the initial
  model, stays frozen between refreshes, and is set to the current parameters when the dense loss
  falls below $\klthresh$. The trained model
  samples rollouts from its last loop $\student$, as in GRPO. The verifier gives one sparse
  advantage $\adv_i$ per response; the teacher gives a dense log-ratio $\logratio_{i,j}$ per
  sampled token. The loss is $\Lgrpo + \kdcoef\,\Lkd$.}
  \label{fig:overview}
\end{figure}

\method keeps GRPO unchanged and adds one dense term that pulls the last loop towards a teacher
built from the loops of an anchor copy of the model, which is frozen between refreshes and replaced
by the current model once the student has absorbed the teacher (\cref{fig:overview}).

\subsection{Preliminaries}\label{sec:prelim}

\paragraph{Looped language models.}
A looped language model applies one shared stack of transformer layers $\nloops$ times to every
token~\citep{universaltransformer,ouro}; Ouro~\citep{ouro} uses $\nloops=4$. With $h^{(t)}(\prefix)$
the normalised hidden state after loop $t$ at the last position of a prefix $\prefix$, the shared LM
head $W$ gives one next-token distribution per loop:
\begin{equation}\label{eq:loopdist}
  \Ploop{t}{\params}(\cdot\mid\prefix) = \softmax\!\big(W\,h^{(t)}(\prefix)\big),
  \qquad t = 1,\dots,\nloops .
\end{equation}
We disable early exit, so the policy is the last loop, $\student \coloneqq \Ploop{\nloops}{\params}$.
We write $\initparams$ for the parameters before RL, $\baseparams$ for the anchor parameters that
build the teacher (\cref{sec:method-refresh}), and
$\lbase{t} \coloneqq \log \Ploop{t}{\baseparams}(\cdot\mid\prefix)$.

\paragraph{GRPO.}
For a prompt $\prompt$, the policy before the update, $\studentold$, samples $\groupsize$ responses
$\resp_i$ with prefixes $\prefix_{i,j} = (\prompt, \resp_{i,<j})$, and a verifier returns
$\reward_i \in \{0,1\}$. GRPO~\citep{grpo} gives every token of response $i$ the group-normalised advantage
\begin{equation}\label{eq:grpo-adv}
  \adv_i = (\reward_i - \operatorname{mean}_{g}\reward)/(\operatorname{std}_{g}\reward + \varepsilon),
\end{equation}
so a group that is all correct or all wrong gives no gradient. With $\mathrm{PPO}_{i,j}(\adv_i)$ the
clipped surrogate of PPO~\citep{ppo} at token $j$, the loss over $\nresp$ responses is
\begin{equation}\label{eq:grpo-loss}
  \textstyle\Lgrpo(\params) = -\frac{1}{\nresp}\sum_{i=1}^{\nresp}\frac{1}{|\resp_i|}
  \sum_{j=1}^{|\resp_i|}\mathrm{PPO}_{i,j}(\adv_i).
\end{equation}
With one optimiser update per step the PPO ratio $\student/\studentold$ is $1$ at the gradient point,
and the per-token gradient is $\adv_i\nabla_{\params}\log\student(\resp_{i,j}\mid\prefix_{i,j})$.

\subsection{A teacher from the model's own loops}\label{sec:method-teacher}

LoopCD~\citep{loopcd} applies contrastive decoding~\citep{cd,dola} inside one looped model, with
the last loop as the strong model and a reference loop $\refloop$ as the weak one: it decodes from
$\log\softmax\big((1+\cdw)\log\Ploop{\nloops}{\params}-\cdw\log\Ploop{\refloop}{\params}\big)$,
$\cdw\ge0$, which in log space extrapolates past the last loop along the direction from loop
$\refloop$ to loop $\nloops$. For Ouro-1.4B-Thinking on math, \citet{loopcd} use $\refloop=1$,
$\cdw=0.5$ and find that most of the gain comes from reordering tokens rather than from sharpening.
Our teacher is this training-free rule applied to the anchor parameters:
\begin{subequations}\label{eq:loopcd}
\begin{equation}\label{eq:teacher}
  \log\teacher(\cdot\mid\prefix)
  = \log\softmax\!\big((1+\cdw)\,\lbase{\nloops} - \cdw\,\lbase{\refloop}\big)
  = \lbase{\nloops} + \cdw\,\loopdir - \log Z_b(\prefix),
\end{equation}
\end{subequations}
where $\loopdir = \lbase{\nloops} - \lbase{\refloop}$ is the anchor's loop direction and
$Z_b(\prefix)$ normalises. Both readouts come from one forward pass of the anchor copy of the same
model, so the teacher needs no second model.

The teacher depends only on $\baseparams$ and $\cdw$ and carries no reward information.
Distillation alone can at best reach the anchor with test-time LoopCD. We therefore keep the
outcome reward and add the teacher as a dense term next to it.

\subsection{Objective}\label{sec:method-objective}

For every sampled token we form the teacher--student log-ratio and the k2 term of KDRL~\citep{kdrl}:
\begin{equation}\label{eq:logratio}
  \logratio_{i,j} = \log\teacher(\resp_{i,j}\mid\prefix_{i,j}) - \log\student(\resp_{i,j}\mid\prefix_{i,j}),
\end{equation}
\begin{equation}\label{eq:kd}
  \Lkd(\params) = \frac{1}{\nresp}\sum_{i=1}^{\nresp}\frac{1}{|\resp_i|}
  \sum_{j=1}^{|\resp_i|}\tfrac12\,\logratio_{i,j}^2 ,
\end{equation}
\begin{equation}\label{eq:objective}
  \Ltotal(\params) = \Lgrpo(\params) + \kdcoef\,\Lkd(\params).
\end{equation}
\begin{sloppypar}
The teacher carries no gradient and is fixed between refreshes (\cref{sec:method-refresh}), so the per-token gradient of \cref{eq:kd} is
$-\logratio\,\nabla_{\params}\log\student(\resp\mid\prefix)$. Since
$\E_{\resp\sim\student}[\nabla_{\params}\log\student] = 0$, its expectation under on-policy sampling is
\[
\E_{\resp\sim\student}\big[(\log(\student/\teacher)+1)\nabla_{\params}\log\student\big]
= \nabla_{\params}\KL{\student(\cdot\mid\prefix)}{\teacher(\cdot\mid\prefix)},
\]
the gradient of the per-token reverse KL at every visited prefix. As in KDRL, each token uses only its own term.
\end{sloppypar}

\paragraph{A dense reward next to the sparse one.}
With the PPO ratio at $1$, the per-token gradient of \cref{eq:objective} is
$-(\adv_i + \kdcoef\,\logratio_{i,j})\nabla_{\params}\log\student(\resp_{i,j}\mid\prefix_{i,j})$,
averaged as in \cref{eq:grpo-loss}. The method is a policy gradient with per-token advantage
$\adv_i + \kdcoef\logratio_{i,j}$. The outcome advantage, constant within a response, splits
probability between correct and incorrect responses. The teacher supplies a per-token reward inside
each response, and it still gives a gradient in groups where $\adv_i = 0$.

\paragraph{Additive, not multiplicative.}
The loop signal enters as a target distribution in an additive term (challenge C2): it is noisy on a
single token, so it is best used as many small
pushes summed over thousands of tokens. A multiplicative design such as
$\adv_i\cdot g(\logratio_{i,j})$ would instead lean on single-token values for credit assignment and
vanish whenever $\adv_i = 0$.

\paragraph{Matched scales.}
Both terms use the same aggregation, a token mean within each response and then a mean over
responses, so $\kdcoef$ means the same for every response length (challenge C3). We use a constant
$\kdcoef$, and $\kdcoef = 0$ recovers the GRPO loss exactly.

\subsection{When to refresh the teacher}\label{sec:method-refresh}

The anchor must move with the student, but not at every step (challenge C1). If the teacher were
recomputed from the current parameters at every step and reached by distillation before each
recomputation, round $k+1$ would target $(1+\cdw)\log\pi^{(k)} - \cdw\,\lbase{\refloop}$, so that,
with loop $\refloop$ held fixed, $\log\pi^{(k)} - \lbase{\refloop} = (1+\cdw)^k\loopdir$ up to
normalisation. The contrast grows geometrically and the iteration has no stable fixed point; the
policy keeps sharpening along $\loopdir$. A teacher that never moves has the opposite problem: once
the student has absorbed it, the dense term only holds the student near the initial model's LoopCD
distribution, and the regularised optimum stays tied to that teacher (\cref{prop:optimum}).

\method keeps the anchor frozen within a phase and refreshes it only when the student has caught up.
The anchor starts at $\baseparams = \initparams$. Every $\refreshint$ steps we average the dense loss
$\Lkd$ of \cref{eq:kd} over those steps; $\Lkd$ is the k2 estimate of the per-token reverse KL from
the student to the teacher, already computed for the gradient. If the average is below a threshold
$\klthresh$, we copy the current parameters into the anchor, $\baseparams \leftarrow \params$, and
the teacher is rebuilt from loops $\refloop$ and $\nloops$ of the new copy; otherwise the anchor
stays. Within a phase the teacher is fixed, so each phase is the KL-regularised problem analysed in
\cref{sec:method-analysis}. Each refresh adds one round of extrapolation, starting from a student
that already matches the previous teacher, so the pace of extrapolation is set by how fast the
student absorbs the teacher rather than by the step count. A refresh also rebuilds loop $\refloop$
from the current model, so the contrast is taken between two loops of the same, current model.

\subsection{What the objective optimises}\label{sec:method-analysis}

We treat each complete response as one choice with its own logit, use exact gradients and assume
sufficient capacity. Up to the group normalisation of \cref{eq:grpo-adv} and the discount-zero
gradient, \cref{eq:objective} estimates the gradient of
\begin{equation}\label{eq:ideal-objective}
  \mathcal{J}(\params)=\E_{\prompt}\,\E_{\resp\sim\student(\cdot\mid\prompt)}\Big[\reward(\prompt,\resp)
  -\kdcoef\sum_{j=1}^{|\resp|}\KL{\student(\cdot\mid\prefix_j)}{\teacher(\cdot\mid\prefix_j)}\Big]
\end{equation}
for the teacher of the current phase. The per-token KL terms add up to one sequence-level KL, so
within a phase this is KL-regularised RL with the teacher as the reference.

\begin{proposition}\label{prop:optimum}
For a fixed prompt $\prompt$, bounded reward and length, and $\kdcoef>0$, the unique maximiser of
\cref{eq:ideal-objective} is
\begin{equation}\label{eq:optimum}
  \pi^\star(\resp\mid\prompt)\propto\teacher(\resp\mid\prompt)\,\exp\big(\reward(\prompt,\resp)/\kdcoef\big).
\end{equation}
For a binary reward, with $p=\teacher(\reward=1\mid\prompt)$ and $q^\star=\pi^\star(\reward=1\mid\prompt)$:
(i) $q^\star=p\,e^{1/\kdcoef}/(p\,e^{1/\kdcoef}+1-p)>p$ if $0<p<1$;
(ii) $\pi^\star\propto\teacher$ inside the correct and inside the incorrect set;
(iii) for any bounded reward, $\E_{\pi^\star}[\reward]$ is non-decreasing in $1/\kdcoef$;
(iv) $H(\pi^\star)\geq q^\star H\big(\teacher(\cdot\mid\reward=1)\big)+(1-q^\star)H\big(\teacher(\cdot\mid\reward=0)\big)$;
(v) $\pi^\star\to\teacher$ as $\kdcoef\to\infty$ and $\pi^\star\to\teacher(\cdot\mid\reward=1)$ as $\kdcoef\to0$ if $p>0$.
\end{proposition}
The proof is in \cref{app:proofs}. The reward thus lifts the student above its teacher, keeps the
teacher's preferences within the correct and within the incorrect responses, and cannot remove all
of the teacher's entropy.

Writing $\teacher\propto\Ploop{\nloops}{\baseparams}e^{\cdw\loopdir}$ splits the KL term at each
prefix as
\begin{equation}\label{eq:decomp}
  -\kdcoef\,\KL{\pi}{\teacher}
  =\kdcoef\,\E_{a\sim\pi}\big[\cdw\loopdir(\prefix,a)-\log Z_b(\prefix)\big]
  -\kdcoef\,\KL{\pi}{\Ploop{\nloops}{\baseparams}},
\end{equation}
following the reading of on-policy distillation as dense RL in ExOPD~\citep{exopd}. The objective
thus combines a sparse outcome reward, a dense per-token reward $\kdcoef\cdw\loopdir$ that favours
tokens the last loop prefers over the first, and a trust region around the anchor's last loop.

Refreshing moves this trust region. In the idealised case $\cdw = 0$, where the teacher is the
anchor's last loop, suppose every phase reaches the optimum \cref{eq:optimum} before the refresh.
Phase $m$ then starts from the teacher $\pi^\star_{m-1}$, and by induction
$\pi^\star_m(\resp\mid\prompt) \propto \Ploop{\nloops}{\initparams}(\resp\mid\prompt)\exp\big((m+1)\,\reward(\prompt,\resp)/\kdcoef\big)$:
$m$ refreshes act like the smaller coefficient $\kdcoef/(m+1)$ while every single phase keeps the
step size of $\kdcoef$. By \cref{prop:optimum}(iii), the expected reward is non-decreasing in $m$;
for a binary reward with $p>0$, (v) gives $\pi^\star_m\to\Ploop{\nloops}{\initparams}(\cdot\mid\reward=1)$
as $m\to\infty$, so a refreshed
anchor no longer caps the student at the initial model's distribution.

\subsection{Implementation}\label{sec:method-impl}

The anchor is a float32 copy of the model, initialised from $\initparams$ and offloaded to CPU
between uses. A refresh copies the current actor parameters into it, and the anchor is saved with
every checkpoint so that training resumes with the same teacher. Each step runs one extra per-loop
forward pass of this copy on the rollouts, reads out loops $\refloop$ and $\nloops$, and stores
$\log\teacher$ of the sampled tokens with the batch. We check for a refresh every
$\refreshint = 5$ steps with threshold $\klthresh = 0.02$.
We use $\cdw = 0.5$ and $\refloop = 1$, the LoopCD setting for Ouro-1.4B-Thinking on
math~\citep{loopcd}, and $\kdcoef = 0.01$. KDRL uses $\kdcoef = 2\times10^{-3}$ with a larger
external teacher; our teacher starts from the student's parameters, so $\logratio$ is small at the
start and we scale $\kdcoef$ up.

\section{Experiments}\label{sec:exp}

\subsection{Setup}\label{sec:exp-setup}

\paragraph{Training.}
All runs start from Ouro-1.4B-Thinking~\citep{ouro}, one stack of 24 layers applied $\nloops = 4$
times per token, and train all parameters on 14,684 problems filtered from
DeepMath~\citep{deepmath} with a binary outcome reward $\reward$. Each step samples 64 prompts and
$\groupsize = 8$ responses per prompt ($\nresp = 512$) from the last loop at temperature 1 and
performs one optimiser update. All runs use 100 steps, learning rate $10^{-6}$, responses of at
most 16,384 tokens, no KL term towards a reference model, and seed 42. \Cref{app:details} lists all
settings.

\paragraph{Compared systems.}
\emph{Base} is Ouro-1.4B-Thinking without further training. \emph{GRPO}~\citep{grpo} trains the last
loop with the clipped PPO surrogate on the outcome advantage. \emph{RLTT}~\citep{rltt} runs the authors' released code
in its original REINFORCE form, which weights the log-probabilities of all loops by the exit distribution.
\emph{OPSD}~\citep{opsd} (on-policy self-distillation) uses the initial model, given privileged
information about the solution, as a fixed teacher for the model without it. On the student's own
rollouts it minimises the per-token generalised Jensen--Shannon divergence (mixing weight $0.5$)
between the two next-token distributions over the full vocabulary, and it uses no outcome reward. \emph{\method} optimises \cref{eq:objective} with the
settings of \cref{sec:method-impl}.

\paragraph{Evaluation.}
We evaluate the step-100 checkpoint on five benchmarks (630 problems). Every
model runs 4 loops, decodes from its last loop without contrastive decoding, and draws 8 samples per
problem (temperature 0.6, top-p 0.95, at most 16,384 tokens, seed 42). We report avg@8, the mean
accuracy over all samples.

\subsection{Main results}\label{sec:exp-main}

Reward-only training gives a small gain (\cref{tab:main}): GRPO raises the average avg@8 from
61.15 to 62.32 ($+1.17$), with changes from $-0.42$ (AIME24) to $+2.92$ (AIME26).

\begin{table}[t]
\centering
\caption{avg@8 on five mathematics benchmarks (Ouro-1.4B-Thinking, 4 loops, 8 samples,
temperature 0.6). Avg.\ is the unweighted mean; bold marks the best method.}
\label{tab:main}
\begin{tabular}{lcccccc}
\toprule
Method & AIME24 & AIME25 & AIME26 & MATH-500 & AMC23 & Avg. \\
\midrule
Base   & 45.42 & 40.83 & 42.08 & 91.78 & 85.63 & 61.15 \\
GRPO   & 45.00 & 40.83 & 45.00 & 92.65 & 88.13 & 62.32 \\
RLTT   & 32.50 & 27.92 & 29.17 & 85.58 & 75.63 & 50.16 \\
OPSD   & 44.58 & 40.00 & 44.58 & 92.20 & 87.50 & 61.77 \\
\method & \textbf{48.33} & \textbf{41.67} & \textbf{47.50} & \textbf{93.20} & \textbf{90.31} & \textbf{64.20} \\
\bottomrule
\end{tabular}
\end{table}

Spreading the reward over loops falls below the base model. RLTT lowers avg@8 on all five
benchmarks, by 6.20 (MATH-500) to 12.92 (AIME24) points, and the average falls to 50.16. \Cref{sec:exp-dynamics} relates the drop to the collapse
of per-loop entropy.

Self-distillation without the reward gives little. OPSD reaches an average of 61.77, $+0.62$ over
the base model and $0.55$ below GRPO, and falls below the base model on AIME24 and AIME25.

Adding the dense teacher term to GRPO gives the best result on every benchmark. \method raises the
average avg@8 to 64.20, $+3.05$ over the base model and $+1.88$ over GRPO, more than doubling the
gain of reward-only training. It is above GRPO on all five benchmarks, by $+0.55$ (MATH-500) to
$+3.33$ (AIME24), and above OPSD on all five by $2.43$ points on average. Its largest gains over
the base model are on AIME26 ($+5.42$) and AMC23 ($+4.68$). The dense term thus adds to the outcome
reward: \method improves on both the reward alone (GRPO) and a dense teacher alone (OPSD).

\subsection{Training dynamics}\label{sec:exp-dynamics}

Training accuracy is the mean reward of the 512 rollouts of a step, and the entropy of loop $t$ is
the token-average entropy of $\Ploop{t}{\params}$ on the rollout tokens, in nats. Because training
accuracy is noisy, we report means over windows of steps.

GRPO keeps the per-loop entropies almost unchanged (loop~1 moves from 1.680 to 1.696 nats between
the first and the last 5 steps). Its training accuracy rises from 0.596 over the first 10 steps to
0.683 over the last 10, and its mean response length falls from 9,563 to 8,520 tokens.

RLTT collapses the entropy of every loop, the earliest loop most. Loop-1 entropy falls from 1.684
to 0.382 nats, 23\% of its start, and is still falling at step 100; all four loops end between 0.36
and 0.41 nats. The collapse buys no accuracy: RLTT's training accuracy is 0.610 over the last 10
steps, against 0.683 for GRPO. The collapse is consistent with training intermediate loops on tokens
sampled from the last loop.

\section{Discussion}\label{sec:conclusion}

\method adds to GRPO a dense per-token term towards a teacher that a looped model already
contains, the cross-loop extrapolated distribution of an anchor copy of itself, refreshed when the student
has absorbed it. Between refreshes the objective is reinforcement learning regularised towards the
teacher, with optimum $\pi^\ast \propto \teacher
\exp(\reward / \kdcoef)$ (\cref{prop:optimum}). On Ouro-1.4B-Thinking it is the best compared
method on all five benchmarks and adds 1.88 points of average avg@8 over GRPO, whereas spreading the
reward over loops falls below the base model.

\method needs one extra forward pass of the anchor copy per step and one CPU copy of the model. Our experiments cover one model
size, one domain (mathematics) and 100 updates. The dense term has discount zero: each token is
rewarded for its own agreement with the teacher, not for its effect on later positions. Natural
next steps are schedules for $\cdw$ and $\kdcoef$, other refresh criteria, a dense term with a non-zero discount, and
larger looped models and other domains.

\bibliographystyle{plainnat}
\bibliography{refs}

\appendix
\section{Proofs and Derivations}\label{app:proofs}

Throughout this appendix we fix a prompt $\prompt$ and the teacher of one phase (between two refreshes), and drop the prompt from the notation; the objective
\cref{eq:ideal-objective} is a sum over prompts, so every statement holds prompt by prompt.
Responses have bounded length (at most 16,384 tokens) and the vocabulary is finite, so the set
$\mathcal{Y}$ of complete responses is finite. A response ends with an end-of-sequence token or at
the length cap, and whether it has ended before position $j$ is determined by $\resp_{<j}$. A policy
$\pi$ assigns $\pi(\resp)=\prod_{j=1}^{|\resp|}\pi(\resp_j\mid\prefix_j)$ with
$\prefix_j=(\prompt,\resp_{<j})$. The teacher $\teacher$ is a softmax at every prefix, so
$\teacher(\resp)>0$ for every $\resp\in\mathcal{Y}$. The reward $\reward$ is bounded. For a binary
reward, $\mathcal{C}=\{\resp:\reward(\resp)=1\}$, $\bar{\mathcal{C}}$ is its complement,
$p=\teacher(\mathcal{C})$, and $\teacher(\cdot\mid\mathcal{C})$ is the teacher conditioned on
$\mathcal{C}$.

\subsection{Proof of \texorpdfstring{\cref{prop:optimum}}{Proposition 1}: sequence-level identity and optimum}

\paragraph{Sum of per-token KL terms.} Let $\mathbb{1}_j=\mathbb{1}[|\resp|\geq j]$, which is a
function of $\resp_{<j}$. Conditional on $\resp_{<j}$, the token $\resp_j$ is distributed as
$\pi(\cdot\mid\prefix_j)$. By the tower property,
\begin{align*}
\E_{\resp\sim\pi}\Big[\sum_{j=1}^{|\resp|}\KL{\pi(\cdot\mid\prefix_j)}{\teacher(\cdot\mid\prefix_j)}\Big]
&=\sum_{j}\E_{\resp\sim\pi}\Big[\mathbb{1}_j\sum_a\pi(a\mid\prefix_j)\log\frac{\pi(a\mid\prefix_j)}{\teacher(a\mid\prefix_j)}\Big]\\
&=\sum_{j}\E_{\resp\sim\pi}\Big[\mathbb{1}_j\log\frac{\pi(\resp_j\mid\prefix_j)}{\teacher(\resp_j\mid\prefix_j)}\Big]\\
&=\E_{\resp\sim\pi}\Big[\log\frac{\pi(\resp)}{\teacher(\resp)}\Big]=\KL{\pi}{\teacher}.
\end{align*}
The third equality uses the autoregressive factorisation of both $\pi(\resp)$ and $\teacher(\resp)$.

\paragraph{Completing the square.} By the identity above,
\[
\mathcal{J}(\pi)=\sum_\resp\pi(\resp)\reward(\resp)-\kdcoef\sum_\resp\pi(\resp)\log\frac{\pi(\resp)}{\teacher(\resp)}.
\]
Define
\[
Z=\sum_\resp\teacher(\resp)e^{\reward(\resp)/\kdcoef},\qquad
\pi^\star(\resp)=\frac{\teacher(\resp)e^{\reward(\resp)/\kdcoef}}{Z},\qquad\text{so that}\qquad
\reward(\resp)=\kdcoef\log\frac{\pi^\star(\resp)}{\teacher(\resp)}+\kdcoef\log Z.
\]
Substituting,
\begin{align*}
\mathcal{J}(\pi)
&=\kdcoef\log Z+\kdcoef\sum_\resp\pi(\resp)\Big[\log\frac{\pi^\star(\resp)}{\teacher(\resp)}-\log\frac{\pi(\resp)}{\teacher(\resp)}\Big]
=\kdcoef\log Z-\kdcoef\,\KL{\pi}{\pi^\star}.
\end{align*}
$Z$ does not depend on $\pi$, and $\KL{\pi}{\pi^\star}\geq0$ with equality if and only if
$\pi=\pi^\star$. Hence $\pi^\star$ is the unique maximiser and $\mathcal{J}^\star=\kdcoef\log Z$.
Since $\pi^\star$ has full support, it can be written autoregressively by taking $\pi^\star(a\mid\prefix)$ as the ratio of the prefix
marginals of $\prefix\circ a$ and $\prefix$, so
it is a valid policy. This is the standard closed form of KL-regularised RL, with the reference
model replaced by $\teacher$.

\paragraph{Part (i).} For a binary reward $Z=p\,e^{1/\kdcoef}+1-p$ and
$q^\star=\sum_{\resp\in\mathcal{C}}\teacher(\resp)e^{1/\kdcoef}/Z=p\,e^{1/\kdcoef}/(p\,e^{1/\kdcoef}+1-p)$.
Equivalently $q^\star/(1-q^\star)=e^{1/\kdcoef}\,p/(1-p)$. For $0<p<1$ and $\kdcoef>0$ the factor
$e^{1/\kdcoef}$ exceeds 1, so $q^\star>p$.

\paragraph{Part (ii).} For $\resp\in\mathcal{C}$,
$\pi^\star(\resp)/\pi^\star(\mathcal{C})=\teacher(\resp)e^{1/\kdcoef}/(p\,e^{1/\kdcoef})=\teacher(\resp\mid\mathcal{C})$,
and in the same way $\pi^\star(\cdot\mid\bar{\mathcal{C}})=\teacher(\cdot\mid\bar{\mathcal{C}})$.
Hence $\pi^\star=q^\star\,\teacher(\cdot\mid\mathcal{C})+(1-q^\star)\,\teacher(\cdot\mid\bar{\mathcal{C}})$.

\paragraph{Part (iii).} Let $\eta=1/\kdcoef$ and $\Lambda(\eta)=\log\sum_\resp\teacher(\resp)e^{\eta\reward(\resp)}$,
the cumulant generating function of $\reward$ under $\teacher$. Writing $\pi^\star_\eta$ for the
optimum at $\kdcoef=1/\eta$, differentiation gives
$\Lambda'(\eta)=\E_{\pi^\star_\eta}[\reward]$ and
$\Lambda''(\eta)=\E_{\pi^\star_\eta}[\reward^2]-\E_{\pi^\star_\eta}[\reward]^2=\Var_{\pi^\star_\eta}(\reward)\geq0$.
So $\E_{\pi^\star_\eta}[\reward]$ is non-decreasing in $\eta$. Because $\pi^\star_\eta$ has the same
support as $\teacher$, the variance is strictly positive, and the increase is strict, whenever
$\reward$ is not constant under $\teacher$; for a binary reward this is $0<p<1$. At $\eta=0$,
$\pi^\star_0=\teacher$ and the expected reward is $\E_{\teacher}[\reward]$.

\paragraph{Part (iv).} By part (ii) the two components of $\pi^\star$ have disjoint supports, so the
entropy splits exactly:
\begin{align*}
H(\pi^\star)
&=-\sum_{\resp\in\mathcal{C}}q^\star\teacher(\resp\mid\mathcal{C})\log\big(q^\star\teacher(\resp\mid\mathcal{C})\big)
 -\sum_{\resp\in\bar{\mathcal{C}}}(1-q^\star)\teacher(\resp\mid\bar{\mathcal{C}})\log\big((1-q^\star)\teacher(\resp\mid\bar{\mathcal{C}})\big)\\
&=h(q^\star)+q^\star H\big(\teacher(\cdot\mid\mathcal{C})\big)+(1-q^\star)H\big(\teacher(\cdot\mid\bar{\mathcal{C}})\big),
\end{align*}
where $h(q)=-q\log q-(1-q)\log(1-q)\geq0$. Dropping $h(q^\star)$ gives the bound in the proposition;
it is also the bound that concavity of entropy gives for any mixture. The fixed point keeps positive
entropy whenever the teacher is not deterministic on the correct set.

\paragraph{Part (v).} As $\kdcoef\to\infty$, $e^{\reward/\kdcoef}\to1$ uniformly because $\reward$ is
bounded, so $Z\to1$ and $\pi^\star\to\teacher$. The objective then reduces to
$\min_\pi\KL{\pi}{\teacher}$, which is pure on-policy distillation. As $\kdcoef\to0$ with a binary
reward and $p>0$, part (i) gives $q^\star\to1$ and part (ii) gives $\pi^\star\to\teacher(\cdot\mid\mathcal{C})$.
Every $\mu$ with $\mu(\mathcal{C})=1$ maximises the expected reward. For such $\mu$,
$\KL{\mu}{\teacher}=\sum_{\resp\in\mathcal{C}}\mu(\resp)\log\frac{\mu(\resp)}{p\,\teacher(\resp\mid\mathcal{C})}
=\KL{\mu}{\teacher(\cdot\mid\mathcal{C})}-\log p\geq-\log p$, with equality only for
$\mu=\teacher(\cdot\mid\mathcal{C})$. So the limit is the reward-optimal policy closest to the
teacher.

\section{Experimental Details}\label{app:details}

\subsection{Hyperparameters}

All trained systems share the settings in \cref{tab:hp-shared}, and \cref{tab:hp-method} lists
the settings that belong to each method.

\begin{table}[h]
\centering
\caption{Settings shared by GRPO, RLTT, OPSD and \method.}
\label{tab:hp-shared}
\small
\begin{tabular}{p{0.36\textwidth}p{0.58\textwidth}}
\toprule
Setting & Value \\
\midrule
Initial model & Ouro-1.4B-Thinking, 24 layers, $\nloops = 4$ loops, full-parameter training \\
Training data & 14,684 filtered DeepMath problems, binary outcome reward \\
Prompts $\times$ responses per step & $64 \times 8 = 512$ \\
Optimiser updates per step & 1 (mini-batch = 64 prompts, gradient accumulation 1, 1 epoch) \\
Training steps & 100 \\
Learning rate & $10^{-6}$, 10 warm-up steps, then constant \\
Optimiser & AdamW, betas $(0.9, 0.99)$, weight decay $0.1$ \\
Gradient clipping & global norm $0.1$ \\
Advantage & group-normalised over the 8 responses of a prompt \\
KL term to a reference model & none (coefficient 0, no reference model loaded) \\
Entropy bonus & none \\
Max prompt / response length & 2,048 / 16,384 tokens \\
Rollout sampling & last loop, temperature 1.0, top-p 1.0, top-k off \\
Seed (training and data order) & 42 \\
Hardware & one node, 8 GPUs \\
\bottomrule
\end{tabular}
\end{table}

\begin{table}[h]
\centering
\caption{Method-specific settings.}
\label{tab:hp-method}
\small
\begin{tabular}{>{\raggedright\arraybackslash}p{0.10\textwidth}>{\raggedright\arraybackslash}p{0.38\textwidth}>{\raggedright\arraybackslash}p{0.42\textwidth}}
\toprule
System & Objective & Method parameters \\
\midrule
GRPO & clipped PPO surrogate on $\adv_i$, last loop & clip $0.2$, dual clip $3$ \\
RLTT & REINFORCE on $\adv_i \sum_t q_t \log \Ploop{t}{\params}$ &
  $q_t$ = exit distribution, exit gate trained \\
OPSD & self-distillation from the initial model given privileged context, no outcome reward &
  fixed teacher, generalised JSD (mixing weight $0.5$), full vocabulary \\
\method & $\Lgrpo + \kdcoef \Lkd$ (\cref{eq:objective}) &
  $\kdcoef = 0.01$, $\cdw = 0.5$, $\refloop = 1$, k2 estimator; refresh check every
  $\refreshint = 5$ steps with $\klthresh = 0.02$ \\
\bottomrule
\end{tabular}
\end{table}

\end{document}